**Motion Capture Analysis of Verb and Adjective Types in Austrian Sign Language**

[1,2,*]Julia Krebs, [3,*]Evie Malaia, [4]Ronnie B. Wilbur, [1,5]Isabella Fessl, [1,5]Hans-Peter Wiesinger, [5]Hermann Schwameder, [1,2]Dietmar Roehm

[1] Department of Linguistics, University of Salzburg, Austria

[2] Centre for Cognitive Neuroscience (CCNS), University of Salzburg, Austria

[3] Department of Communicative Disorders, University of Alabama, Tuscaloosa, USA

[4] Department of Linguistics, and Department of Speech, Language, and Hearing Sciences, Purdue University, West Lafayette, Indiana, USA

[5] Department of Sport and Exercise Science, University of Salzburg, Austria

*These two authors contributed equally to this work

**Abstract**

Across a number of sign languages, temporal and spatial characteristics of dominant hand articulation are used to express semantic and grammatical features. In this study of Austrian Sign Language (Österreichische Gebärdensprache, or ÖGS), motion capture data of four Deaf signers is used to quantitatively characterize the kinematic parameters of sign production in verbs and adjectives. We investigate (1) the difference in production between verbs involving a natural endpoint (telic verbs; e.g. arrive) and verbs lacking an endpoint (atelic verbs; e.g. analyze), and (2) adjective signs in intensified vs. non-intensified (plain) forms. Motion capture data analysis using linear-mixed effects models (LME) indicates that both the endpoint marking in verbs, as well as marking of intensification in adjectives, are expressed by movement modulation in ÖGS. While the semantic distinction between verb types (telic/atelic) is marked by higher peak velocity and shorter duration for telic signs compared to atelic ones, the grammatical distinction (intensification) in adjectives is expressed by longer duration for intensified compared to non-intensified adjectives. The observed individual differences of signers might be interpreted as personal signing style.

## 1. Introduction

Previous work has shown that across sign languages, perceptually available dynamics of movement in events and activities are recruited to reflect distinctions in meaning and grammar (Strickland et al., 2015). For example, in American Sign Language (ASL), the verbs that denote telic events (those with an inherent end-point, e.g. arrive) have sharper, more abrupt articulation as compared to the verbs that denote atelic events (those without an inherent end-point, denoting states or activities, e.g. analyze) (Wilbur, 2003). Comparisons of unrelated sign languages, such as ASL and Croatian Sign Language (HZJ), revealed cross-linguistic differences in the strategy of recruiting perceptually available properties of motion for representation of event types (telic/atelic distinction). For example, in ASL, telic verbs are articulated with greater deceleration to a stop at the end of the sign than atelic verbs (Malaia and Wilbur, 2012). In HZJ, the signs for a number of telic verbs can be derived from atelic sign roots by way of changes in the motion of the dominant hand (Milković, 2011). These derived telic signs are distinguished from atelic root signs by greater deceleration and peak velocity of the dominant hand, as confirmed by motion capture data (Malaia et al., 2013).

The strategy of recruitment of physical properties of motion for grammatical marking in sign languages goes beyond verbs: for example, intensified ASL adjectives (e.g. "very heavy") also speeded articulation of the endpoint marking (Wilbur et al., 2012). Hearing non-signers also perceive such motion-based distinctions in sign production, and extrapolate the meaning behind motion to aspectual and event structure features of spoken languages known to them (Strickland et al., 2015; Malaia et al., 2012; Krebs et al., 2023). However, quantitative data on the recruitment of articulation kinematics to mark grammatical functions in sign languages is still sparse.

This study uses motion capture to investigate the kinematic features in production of verbs and adjectives in Austrian Sign Language (ÖGS), focusing on the marking of 1) the telic/atelic distinction between verb signs, and 2) intensification in adjective pairs (cf. cold - very cold). The roots of telic and atelic verb signs in ÖGS are unrelated phonologically (i.e. with regard to handshape, hand orientation, and place of articulation), while intensified adjective forms are derived from plain (non-intensified) forms. Thus, we are interested in identifying the kinematic characteristics of signed grammar in comparing a semantic distinction (telic/atelic) to a grammatical distinction (intensification derivation) in ÖGS. It was hypothesized that grammatical markers for verbs and adjectives are generated based on physical properties of articulator motion in space (i.e. velocity and acceleration), as has been attested for other sign languages (ASL and HZJ). The specifics of physical grammatical markers could not be hypothesized, as it is the first study of this kind on ÖGS.

## 2. Materials and methods

### 2.1. Participants

Four Deaf signers (2 F) were included in the analysis (Age M=54, SD = 10, range 40-64). All participants were born deaf or lost their hearing early in life. All of the participants who took part in the study were fluent ÖGS signers, used ÖGS as their first language in daily life, are members of the Deaf community, and have been associated with our research for many years. Three participants self-reported as right-handed; one self-reported as left-handed.

### 2.2. Materials and design

The list of signs each participant produced consisted of 102 signs. The stimuli included 36 telic and 36 atelic verb signs (e.g. telic: arrive; atelic: write), 15 adjectives in non-intensified form (e.g. sweet), and the same 15 adjectives in intensified form (e.g. very sweet). Stimuli were presented in a power point presentation, a written gloss of each sign on a separate slide. The stimuli were elicited in pseudo-randomized order, such that no sign type appeared more than two times in a row. Every other participant was presented with the list in the reversed order to eliminate potential order effects.

### 2.3. Data collection and analysis

Body kinematics of the trunk, head, and upper extremities including hands were recorded using a custom-built marker set (see Fig.1), and a 12-camera infrared motion capture system (Qualisys AB, Göteborg, Sweden) with a sampling frequency of 300 Hz. A 2D-Video (150 Hz, Qualisys AB, Göteborg, Sweden) of the participant's frontal plane was recorded simultaneously, and time-locked to motion capture data.

Marker trajectories were low-pass filtered using a second-order, zero-lag Butterworth filter with a cutoff frequency of 25 Hz. Segment positions and orientations were determined using an inverse kinematics algorithm (V3D; C-Motion, Rockville, MD, USA). Joint centers of the wrist, elbow, and shoulder were defined as virtual landmarks at 50% of the line between the lateral and medial joint markers. The velocity of the wrist joint center (vertical component) of the dominant hand was used to define the onset ($v > 0.1$ m/s) and offset ($v < 0.1$ m/s) of hand movement.

The dominant hand in sign language production is the one that is used for signing one-handed signs. In two-handed asymmetric signs (i.e. the two hands show different movement and different handshape) the dominant hand executes the primary movement and the second (non-dominant) hand functions as a ground, or place of articulation, or replicates the primary movement. For two of the signers who self-reported being right-handed, the dominant hand was always the right hand. The signer who self-reported being left-hand-dominant used the left hand as the dominant one, with the exception of the sign write; finally, one signer alternated in using the right or the left hand as the dominant hand. For statistical analysis, each sign was evaluated individually, and the dominant hand data for each signer and sign was used.

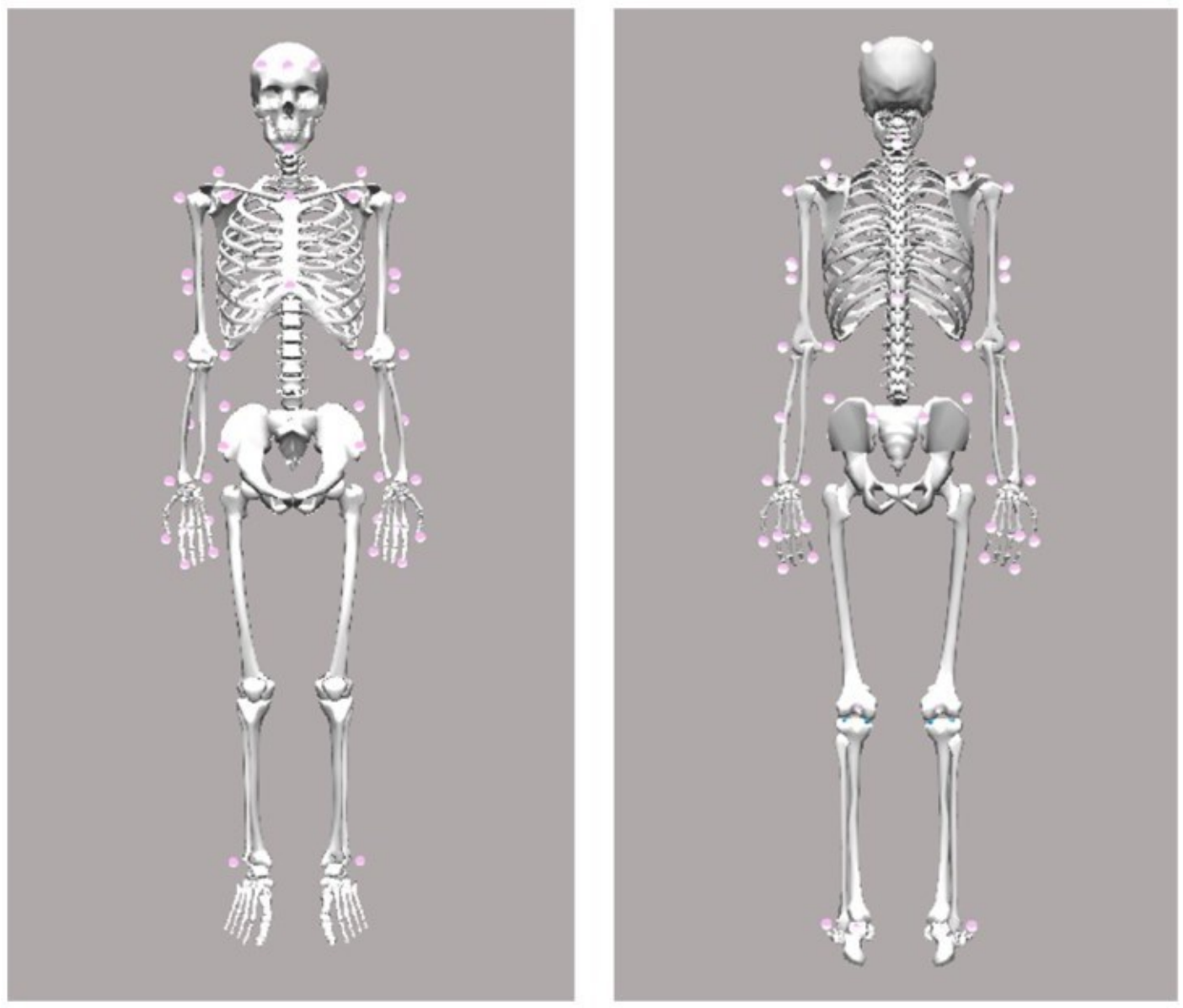

**Figure 1: Marker set on skeletal representation.**

The start and end of the sign phase was set visually by a skilled signer using 2D video recording time-aligned to motion capture data. Sign onset was defined as the video frame when the target handshape reached target location from where sign movement started (Wilbur and Malaia, 2008). Sign offset was defined as the video frame when the hand changed its shape or orientation or moved away from the final position. The complete sign was divided into 3 phases: preparation phase (hand movement onset – start sign), sign phase (start sign – end sign), and down phase (end sign – hand movement offset). Resultant absolute mean and peak velocity and acceleration of the joint centers were calculated for all three sign phases; the present analysis focused on sign phase exclusively.

2.4. Statistical analysis

2.4.1. Verbs

The effect of Verb type was examined separately for two kinematic dependent variables: (a) peak velocity and (b) verb duration (both were log-transformed). The statistical analyses were conducted using linear mixed-effects (LME) models, and performed using the lme4 package (Bates et al., 2015) in R (R Core Team, 2018), with Verb type (telic vs. atelic) as a fixed effect. The random effects included by-participant and by-item random intercepts. Models with random slope for Verb type in the by-participants or by-item term were also tested for convergence. Sum coding was used for main effects testing in all models. For peak velocity as an outcome variable, only the model with by-participant and by-item random intercepts converged[1]. Among the models with duration as an outcome variable, the model with additional random slope for Verb type[2] in the by-participants term provided the better fit for the data (i.e. lower AIC/BIC values)

[1] coded in R as lme = lmer(log(Peak Velocity+1) Verbtype + (1|Participant) + (1|Item))

[2] coded in R as lme = lmer(log(Duration+1) Verbtype + (1+Verbtype|Participant) + (1|Item))

among the converged models. A t-value of 2 and above was interpreted as indicating a significant effect (Baayen et al., 2008); p-values were assessed using the lmerTest package using maximum likelihood estimators (MLE).

2.4.2. Adjectives

Statistical analysis for kinematic features of adjectives was computed using LME modeling similar to the analysis for verb kinematics reported in the previous section. The fixed effect of Intensification (non intensified vs. intensified) was examined separately for two dependent variables: (a) peak velocity and (b) adjective sign duration. For peak velocity, only the model with by-participant and by-item random intercepts converged[3]. For duration, the model with by-participant and by-item random intercepts is reported[4]; the model with additional random slope for Intensification in the by-item term converged, but did not provide a better fit for the data based on AIC/BIC.

## 3. Results

3.1. Verbs

3.1.1. Peak velocity

Table 1 provides an overview of average peak velocity in atelic and telic verb signs. Atelic verbs were signed with a lower peak velocity compared to telic ones (cf. Fig. 2). The mixed-effects model for peak velocity revealed an effect of Verb type (Estimate: -.076; SE: .017; p < .001).

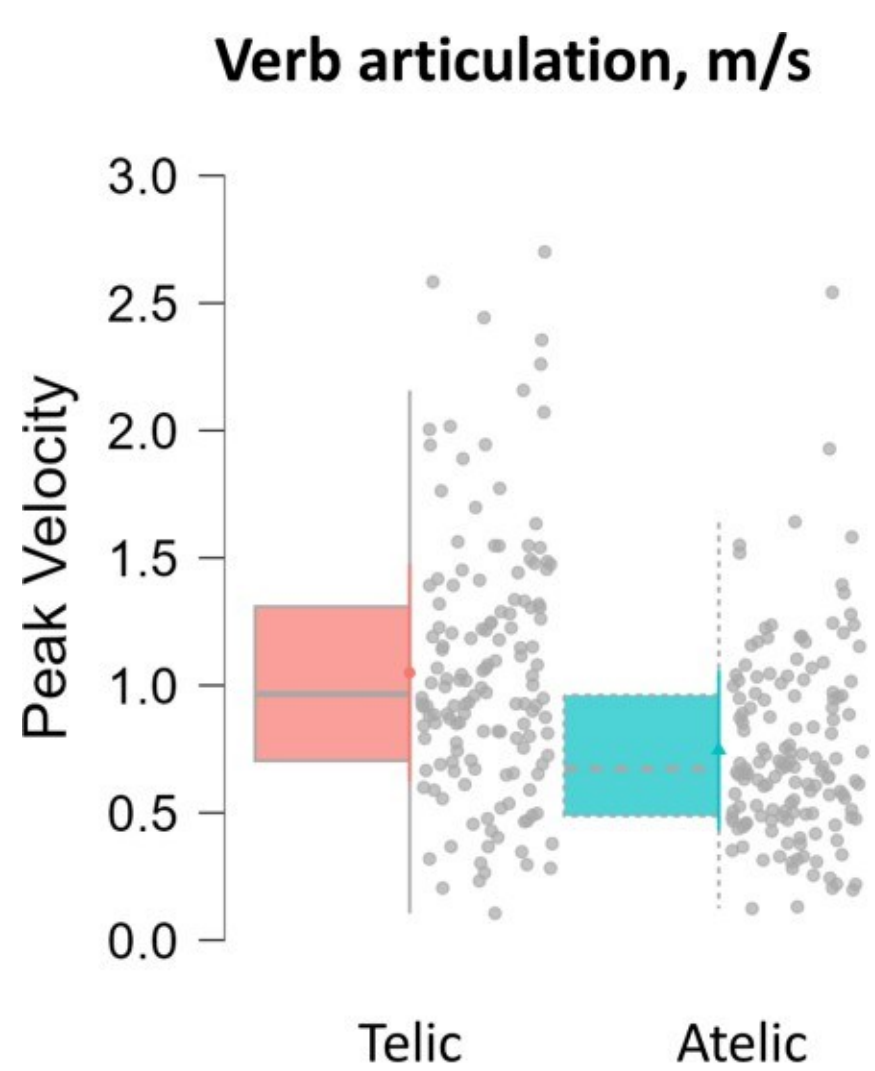

**Figure 2: Verb peak velocity comparison.**

[3] coded in R as lme = lmer(log(Peak Velocity+1) Intensification + (1|Participant) + (1|Item))
[4] coded in R as lme = lmer(log(Duration+1) Intensification + (1|Participant) + (1|Item))

3.1.2. Duration

Table 1 provides an overview of average duration in atelic and telic verb signs. Atelic verbs were longer in duration compared to telic ones (see Fig. 3). The mixed-effects model for duration revealed an effect of Verb type (Estimate: .101; SE: .023; p < .01).

| Variable | Telic | Atelic |
| --- | --- | --- |
| Peak velocity, m/s | 1.05 (0.50) | 0.75 (0.37) |
| Duration, s | 0.95 (0.25) | 1.39 (0.40) |

Table 1: Mean peak velocity and duration of telic and atelic verbs. Standard deviation is presented in parentheses.

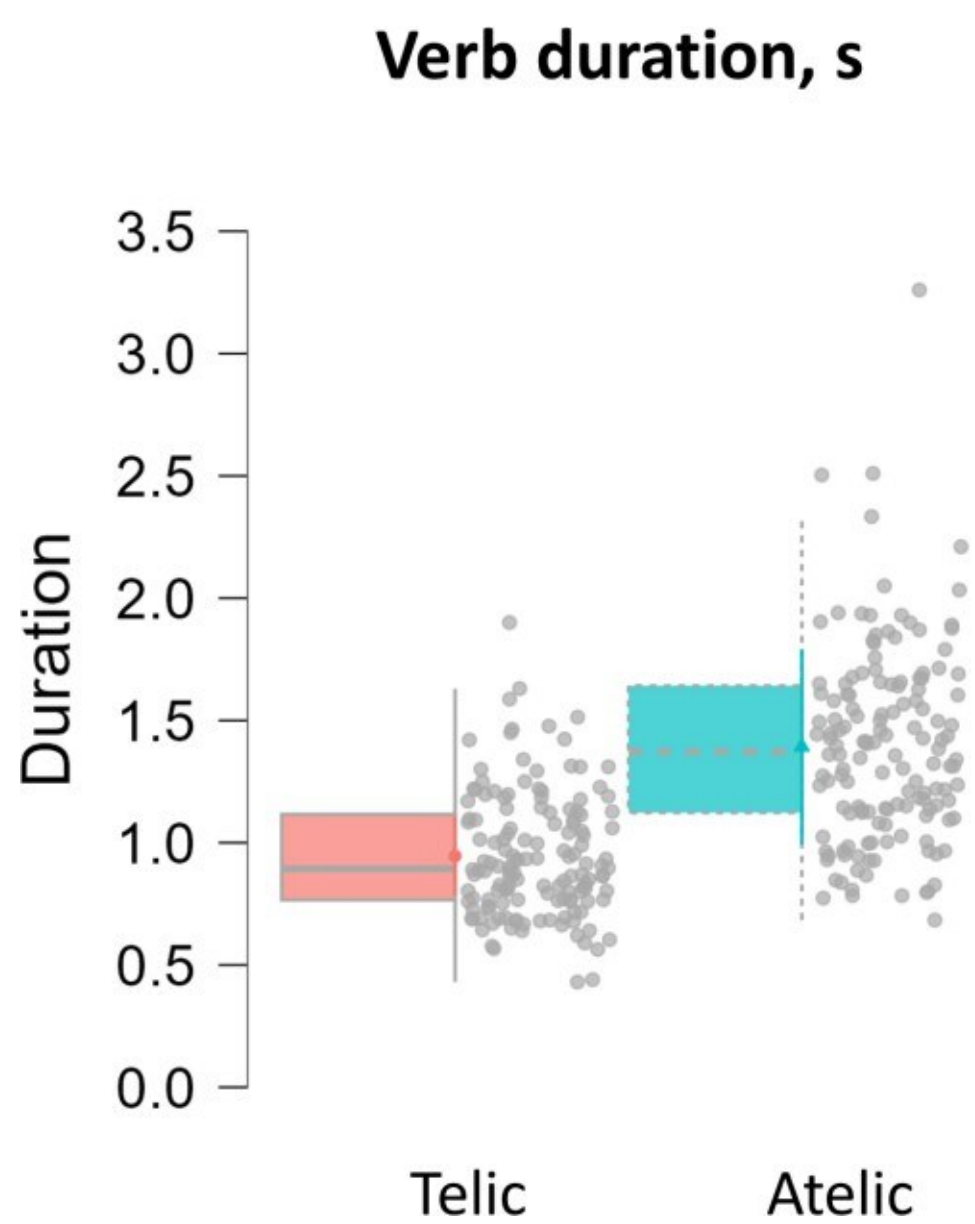

**Figure 3: Verb duration comparison.**

3.2. Adjectives

3.2.1. Peak velocity

Table 2 provides an overview of average peak velocity in non-intensified and intensified adjectives. Intensified adjectives were signed with a higher peak velocity compared to non-intensified adjectives (see Fig. 4). However, the mixed-effects model for peak velocity did not reveal an effect of Intensification (Estimate: .032; SE: .025; p = .22).

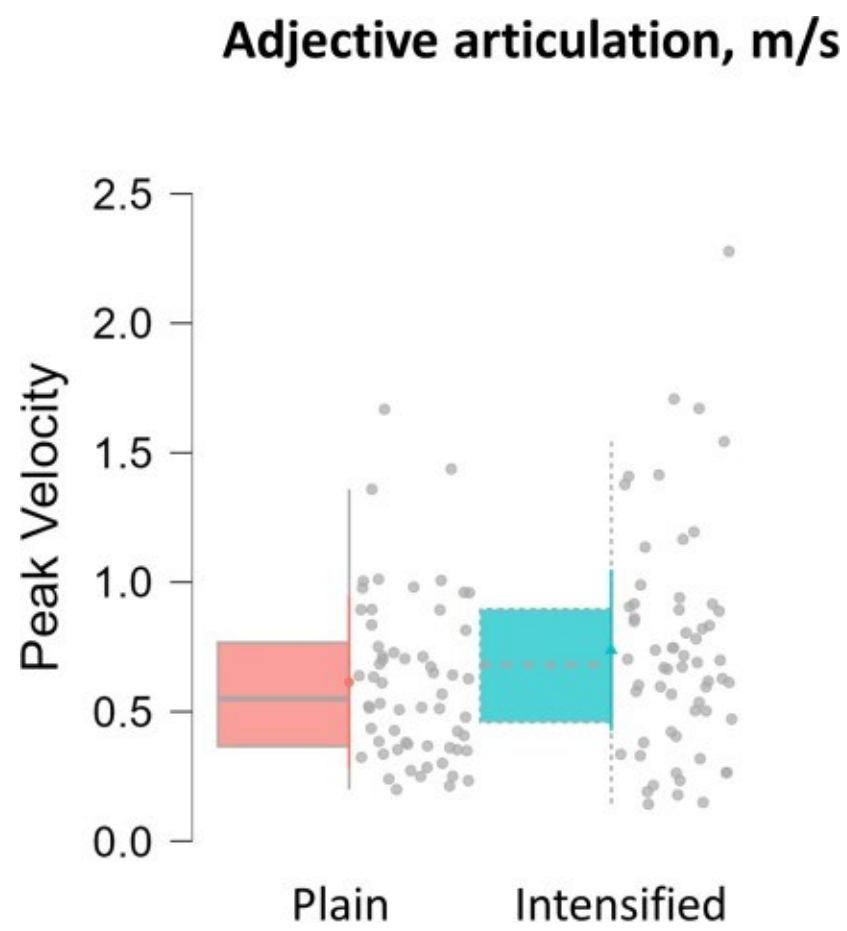

**Figure 4:** Adjective velocity comparison.

3.2.2. Duration

Table 2 provides an overview of average duration of non-intensified and intensified adjectives. Intensified adjectives are longer in duration compared to non-intensified adjectives (Fig. 5). The mixed-effects model for duration revealed an effect of Intensification (Estimate: .050; SE: .024; $p < .05$).

| Variable | Plain | Intensified |
|---|---|---|
| Peak velocity, m/s | 0.61 (0.31) | 0.74 (0.43) ) |
| Duration, s | 1.01 (0.34) | 1.24 (0.53) |

Table 2: Mean peak velocity and duration of intensified and non-intensified adjectives. Standard deviation is presented in parentheses.

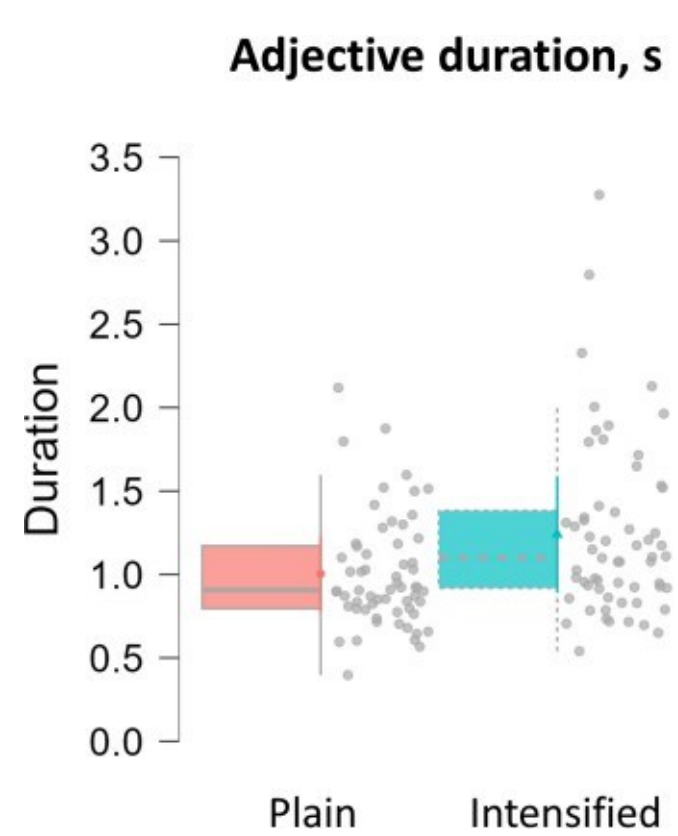

**Figure 5:** Adjective duration comparison.

## 4. Discussion

The ÖGS data indicates that telic verb signs were produced with a higher peak velocity and shorter in duration as compared to atelic verb signs. Thus, the linguistic difference in semantics (telic-atelic distinction) appears associated with kinematic differences in both peak velocity and duration in ÖGS, corroborating an earlier pilot analysis with one signer (not included in this study) (Krebs et al., 2021).

Intensified adjectives were longer in duration compared to non-intensified adjectives. The grammatical marking of intensification appears to be associated primarily with differences in duration. Duration differences were observed both between verb types (telics vs. atelics) and between adjective types (intensified vs. non-intensified); however, the difference in duration may stem from different kinematic properties in the types of stimuli. The longer duration in atelic verbs seems to be related to the phonological structure of the signs (i.e. most of them show a reduplicated movement component leading to a longer sign duration). The duration in adjectives, however, seems to be connected to both the span of signing space (size of the sign) and velocity of hand motion, but not to sign repetition.

The data also revealed individual differences among the signers for both verbs (Fig. 6) and adjectives (Fig. 7), which might be interpreted as personal signing style (cf. Bigand et al., 2020).

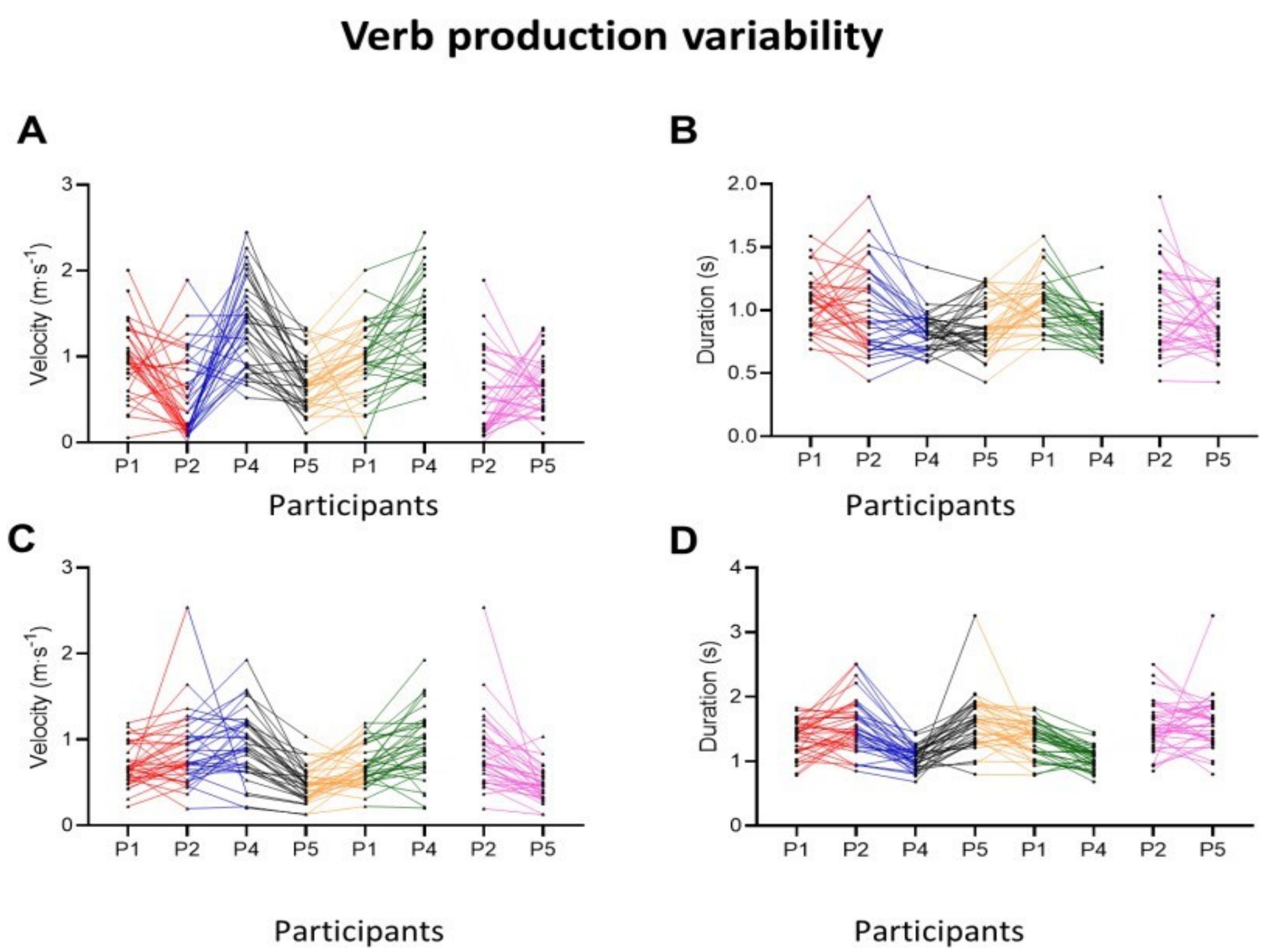

**Figure 6:** Pairwise comparison of 4 fluent Deaf signers (P1-P4) dominant wrist joint center velocity and duration variability within the sign phase for telic verbs (A, B) and atelic verbs (C, D).

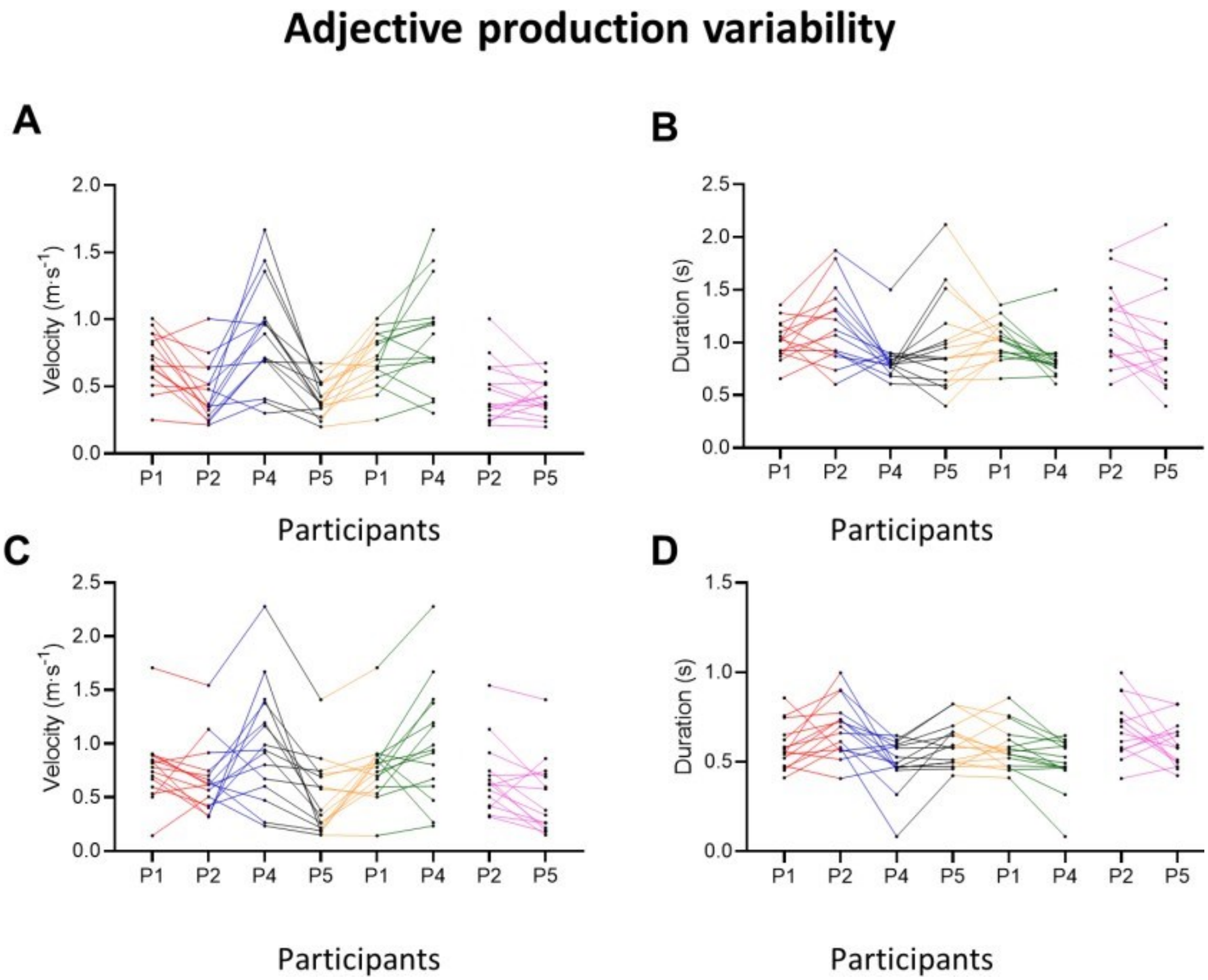

**Figure 7:** Pairwise comparison of 4 fluent Deaf signers (P1-P4) dominant wrist joint center velocity and duration variability within the sign phase for plain adjectives (A, B) and intensified adjectives (C, D).

The present study suggests that in sign languages the physical parameters of motion are recruited for semantic and grammatical markings, but different parameters are recruited for different marking categories. The findings also point to cross-linguistic differences: all languages in which kinematic properties have been investigated so far in the context of verbs and adjectives use duration, velocity, and acceleration as grammatical markers, but weigh the salience of each physical marker differently. More participant data is needed to further investigate the hypothesis tested in the present study.

Research on the dynamic characteristics of the sign language signal is important for future development of applications such as improved automatic recognition/translation of signed material, construction of animated avatar models, sign language teaching, interpreter training, and educational materials.

**Acknowledgements**

We want to thank our participants for taking part in this study. This work was supported in part by the Austrian Science Fund (FWF): 35671, the Austrian Science Fund (FWF): ESP 252-G and by the National Science Foundation (NSF) award #1932547.